\documentclass[10pt,twocolumn,letterpaper]{article}
\usepackage{iftex}
\ifPDFTeX
  \pdfoutput=1
\fi

\PassOptionsToPackage{table}{xcolor}
\usepackage[pagenumbers]{cvpr}
\usepackage[T1]{fontenc}
\usepackage{multirow,makecell}

\definecolor{cvprblue}{rgb}{0.21,0.49,0.74}
\usepackage[breaklinks,colorlinks,allcolors=cvprblue]{hyperref}
\definecolor{prismopen}{RGB}{255,242,224}
\definecolor{prismclosed}{RGB}{253,234,237}
\definecolor{prismreward}{RGB}{240,243,246}
\definecolor{prismours}{RGB}{227,242,253}
\newcommand{\prismdelta}[1]{\textcolor{green!60!black}{\textbf{#1}}}

\makeatletter
\patchcmd{\@maketitle}{\vspace*{24pt}}{\vspace*{8pt}}{}{%
  \PackageError{PreserveReward}{Could not adjust title-author spacing}{Check the CVPR title macro.}}
\patchcmd{\@maketitle}{\vspace*{12pt}}{\vspace*{6pt}}{}{%
  \PackageError{PreserveReward}{Could not adjust author-body spacing}{Check the CVPR title macro.}}
\makeatother

\title{PreserveReward: Reliable Consistency Reward Modeling for Image Editing via Dual-Teacher On-Policy Self-Distillation}
\author{%
Xueheng Li$^{1,2}$\thanks{Equal contribution.}\quad Yong Liu$^1$\footnotemark[1]\quad Wen Xue$^{1,3}$\quad Shi-Yu Tian$^{1,4}$\\
Zhuofan Xia$^1$\quad Chengjun Xie$^2$\quad Yipeng Sun$^1$\thanks{Corresponding author.}\quad Simiu Gu$^1$\\[0.4em]
{\large $^1$JD.com}\\
{\large $^2$University of Science and Technology of China}\\
{\large $^3$South China University of Technology\quad $^4$Nanjing University}
}
\hypersetup{%
  pdftitle={PreserveReward: Reliable Consistency Reward Modeling for Image Editing via Dual-Teacher On-Policy Self-Distillation},
  pdfauthor={Xueheng Li, Yong Liu, Wen Xue, Shi-Yu Tian, Zhuofan Xia, Chengjun Xie, Yipeng Sun, Simiu Gu},
  pdfsubject={Image editing consistency reward modeling},
  pdfkeywords={image editing, reward modeling, on-policy self-distillation, content preservation}
}

\begin{document}
\raggedbottom
\makeatletter
\AddToHookNext{shipout/after}{\global\let\@textbottom\relax}
\makeatother
\maketitle

\begin{abstract}
Instruction-guided image editing has a wide range of applications, yet existing models may inadvertently alter content that should remain unchanged. Reliable consistency evaluation is therefore essential for assessing content preservation and providing reward signals for Reinforcement Learning (RL). However, existing reward models may overlook subtle visual changes or fail to distinguish necessary modifications from unintended alterations, limiting the reliability of consistency assessment. To address this challenge, we propose PreserveReward, a consistency reward model for image editing that integrates visual On-Policy Self-Distillation (OPSD) to enhance fine-grained visual perception and consistency assessment. PreserveReward learns from reference preservation scores through a regression-based scoring loss and employs two complementary teacher views for self-distillation. The local teacher uses matched crops to identify subtle visual changes, while the joint teacher combines local details with global context to determine whether these changes align with the editing intent. To improve the reliability of distillation supervision, we calibrate teacher score distributions via minimum-KL projection under reference-score error constraints and adaptively select teacher process supervision based on gradient alignment with the scoring objective, mitigating scoring bias and optimization conflicts. Extensive experiments across multiple benchmarks demonstrate that PreserveReward achieves strong alignment with human preference on non-edited content preservation and generalizes well across diverse editing scenarios.

\end{abstract}

\section{Introduction}
\label{sec:intro}
\begin{figure}[ht]
    \centering
    \includegraphics[width=\linewidth]{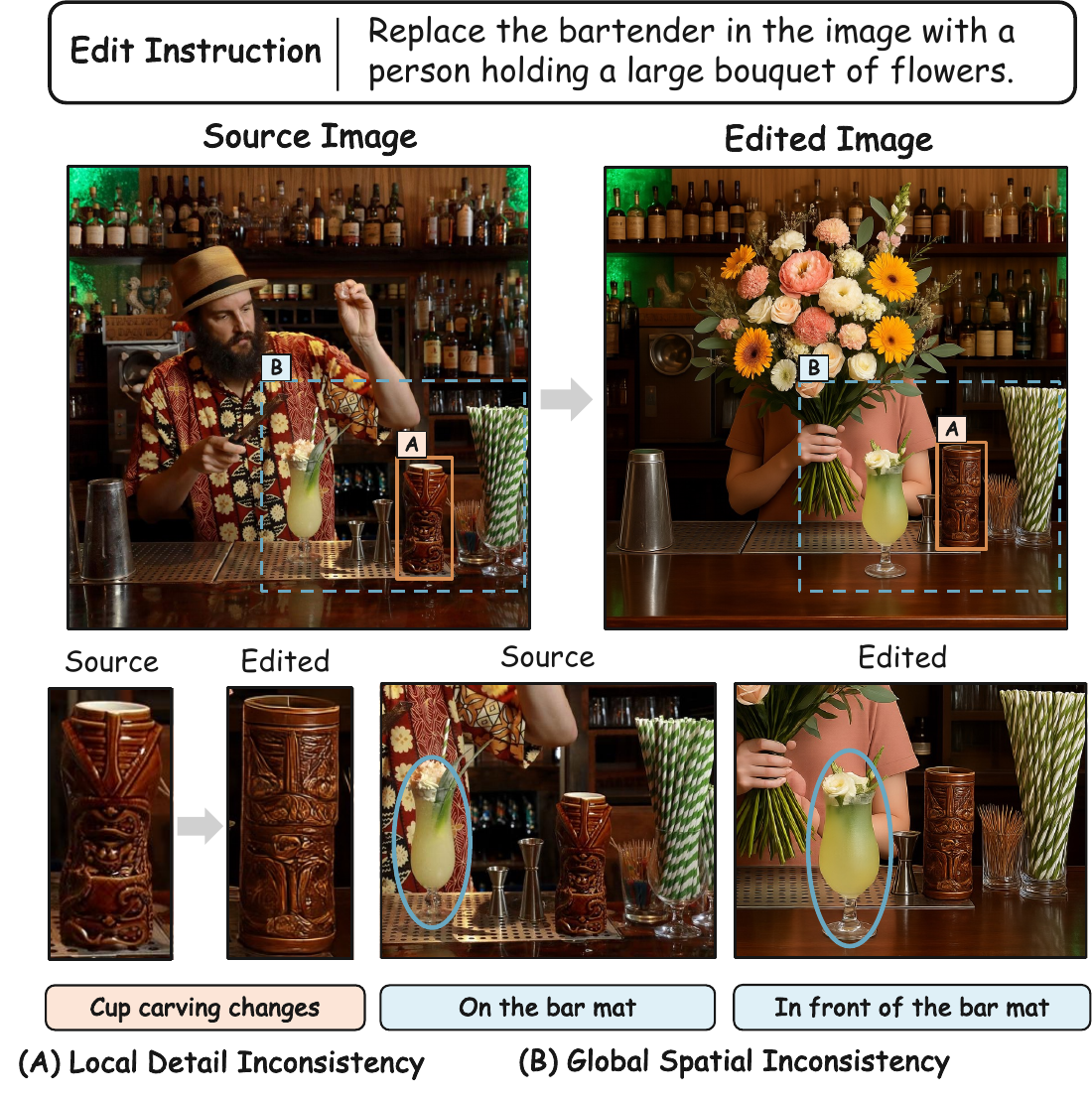}
    \caption{Examples of non-edited content inconsistencies: (A) subtle changes in local object details and (B) unintended changes in spatial relationships.}
    \label{fig:motivation}
\end{figure}    
Instruction-guided image editing supports a wide range of applications, from photo retouching to creative design~\cite{zhou2025multi}. Although recent models have made substantial progress in modifying target objects according to user instructions~\cite{wu2025qwenimagetechnicalreport}, they can still inadvertently alter objects that should remain unchanged~\cite{sahay2026sourcereward}. This limitation calls for reward models that accurately assess non-edited subject consistency, enabling reliable evaluation of content preservation and providing reward signals for Reinforcement Learning (RL) to improve editing fidelity~\cite{luo2026editscore}. Developing such models is challenging because consistency may depend on fine-grained object attributes or spatial relationships across the scene~\cite{zhao2026trust}. An evaluator must detect subtle changes in these properties and interpret them in light of the editing instruction, drawing on global context when needed to distinguish necessary modifications from unintended alterations~\cite{kim2025preserve}.

Guiding policy optimization toward instruction following and content preservation requires reliable reward signals. EditReward~\cite{wu2025editreward} learns to score edits from human preferences, while EditScore~\cite{luo2026editscore} generates reasoning before scoring to support online RL. FIRM~\cite{zhao2026trust} explicitly assesses edit execution and consistency separately. Despite these advances, reliable consistency evaluation remains challenging. Evaluators emphasizing global semantics may overlook subtle changes to the markings, textures, or identity cues of non-edited subjects, underestimating the resulting loss of fidelity~\cite{zhao2026trust,xu2026qwen}. Meanwhile, consistency also encompasses object counts, spatial relationships, and scene layout, requiring a joint assessment of local details and global structure, as illustrated in Fig.~\ref{fig:motivation}. Crucially, comparing source and edited images also requires interpreting these changes in light of the editing instruction and scene context to distinguish necessary modifications from unintended alterations. Without complementary evidence across scales, reward models may overlook damage to content that should remain unchanged or incorrectly penalize intended edits, producing unreliable optimization signals. These challenges persist even in advanced open-source vision-language models (VLMs), which can exhibit substantial errors in assessing content preservation; increasing model size alone does not necessarily ensure reliable consistency judgments~\cite{luo2026editscore,qwen2026qwen38_27b}. Although proprietary VLMs can offer more accurate assessments, the cost and latency of repeated reward queries constrain their practical use in online RL. These limitations motivate consistency reward models that integrate fine-grained visual perception with global contextual reasoning, improve scoring reliability, and maintain efficient inference for large-scale policy optimization.

To address these challenges, we propose PreserveReward, a consistency reward model for image editing that builds on On-Policy Self-Distillation (OPSD) to improve fine-grained perception of non-edited visual content and scoring reliability. Given an editing instruction and the source and edited images, the model generates a consistency rationale and a numerical score. We supervise score predictions with a Score-MSE loss, which weights each candidate score's squared error relative to the reference target by the probability assigned to that score. However, score supervision alone provides limited guidance on identifying and interpreting visual evidence relevant to consistency. We therefore introduce two complementary teacher views, local and joint, to enhance the student's visual perception and reasoning through self-distillation. The local teacher examines matched crops from the source and edited images to identify subtle changes in subject identity and texture. The joint teacher additionally observes the full image pair, considering object counts, spatial relationships, and scene layout to assess whether local changes are consistent with the editing intent. The teachers supervise both reasoning and score prediction along student-generated trajectories, transferring the benefits of additional visual evidence to the student's full-image inference.

To leverage complementary visual evidence for reliable consistency reward modeling, we improve teacher supervision in two aspects: optimization compatibility and scoring reliability. First, to mitigate potential optimization conflicts between teacher distillation and reference-score supervision, we introduce score-guided gradient routing. For each training batch, this strategy selects compatible teacher feedback based on the alignment between each teacher's combined distillation gradient and the Score-MSE gradient, and skips distillation updates when both alignments are negative. Meanwhile, although additional visual evidence enables more fine-grained consistency judgments, teacher numerical scores may still deviate from the reference scores, and directly distilling teacher predictions may propagate scoring bias. We therefore calibrate both teachers' score distributions through a minimum-KL projection under an expected scoring-error constraint defined by the reference scores. This controls expected scoring error while minimally modifying the original teacher distributions, yielding calibrated targets for score distillation. Together, these mechanisms help better leverage complementary visual evidence to produce more reliable consistency rewards. PreserveReward uses additional local visual evidence only during training and requires only the editing instruction and full image pair at inference. Extensive experiments across multiple datasets demonstrate the effectiveness of PreserveReward for image editing consistency evaluation, with strong scoring accuracy and generalization.

In summary, our contributions are as follows:
\begin{itemize}
    \item We propose PreserveReward, an image editing consistency reward model based on on-policy self-distillation, designed to enhance fine-grained perception of non-edited content and improve scoring reliability.
    \item We introduce a score-constrained teacher distribution calibration strategy that employs minimum-KL projection under reference-score error constraints to mitigate teacher scoring bias and provide more reliable distillation supervision.
    \item We develop a dual-teacher self-distillation and gradient routing mechanism that integrates complementary local and global visual evidence and adaptively selects teacher process supervision based on score-supervision feedback, reducing optimization conflicts between distillation and score learning.
    \item Extensive experiments across multiple datasets demonstrate the effectiveness of PreserveReward for image editing consistency evaluation.
\end{itemize}

\section{Related Work}
\subsection{Reward Models for Image Editing}
Reward models for instruction-guided image editing evaluate a candidate edit given the source image and the editing instruction. HIVE~\cite{zhang2024hive} learns a scalar reward from human preferences to fine-tune diffusion editors, while EditReward~\cite{wu2025editreward} scales up preference supervision with expert-annotated comparisons and uses the learned scorer to filter synthetic training data. EditScore~\cite{luo2026editscore} uses a generative evaluator to assess semantic consistency and perceptual quality, supporting both candidate selection and online reinforcement learning. FIRM~\cite{zhao2026trust} uses descriptions of source-to-edit differences to construct supervision for separate execution and consistency scores. EditHF~\cite{xu2026edithf1mmillionscalerichhuman} combines score prediction with pairwise comparison and evaluates attribute preservation alongside instruction alignment and perceptual quality. SpatialReward~\cite{long2026spatialreward} predicts edit regions to guide comparisons between the source and edited images. HP-Edit~\cite{li2026hp} trains a vision--language evaluator on human preference scores and uses it to construct preference data and provide rewards for editor post-training. SourceReward~\cite{sahay2026sourcereward} learns from preferences aggregated across multiple judges and uses edit-centric cross-attention to assess instruction alignment and source preservation. However, existing evaluators emphasizing global semantic consistency may overlook subtle changes to non-edited subjects, such as alterations in markings, textures, or identity cues, leading to unreliable consistency rewards. 

\subsection{Visual On-Policy Self-Distillation}
On-policy self-distillation transfers teacher supervision along trajectories sampled from the student policy. Recent work has extended this approach from language model post-training to multimodal settings, where visual evidence can guide teacher supervision. For image understanding, Vision-OPD~\cite{yuan2026vision} uses evidence-centered crops to supervise a full-image student, and ViCuR~\cite{tian2026vicur} uses query-relevant visual cues as training-time privilege. Contrasting visual conditions also provides a way to control which signals are distilled. VCSD~\cite{liang2026visual} contrasts teacher predictions on the original and content-erased images to reshape the distillation target. OPD-V~\cite{bi2026opd} uses the prediction margin between zoomed and masked teacher views to select and weight token-level supervision, while EAD~\cite{hu2026evidence} weights teacher corrections using the student's response to evidence masking. V-Zero~\cite{sun2026v} uses contrastive visual evidence to gate distillation at the trajectory level. For long-video understanding, Clue-OPSD~\cite{wang2026look} conditions the teacher on annotated clue intervals to supervise a full-video student. However, applying visual OPSD to consistency reward modeling requires aligning teacher scores with human judgments. Direct distillation may propagate scoring bias and introduce optimization conflicts with reference-score supervision, undermining scoring reliability.

\section{Method}

\begin{figure*}[t]
    \centering
    \includegraphics[width=0.9\textwidth]{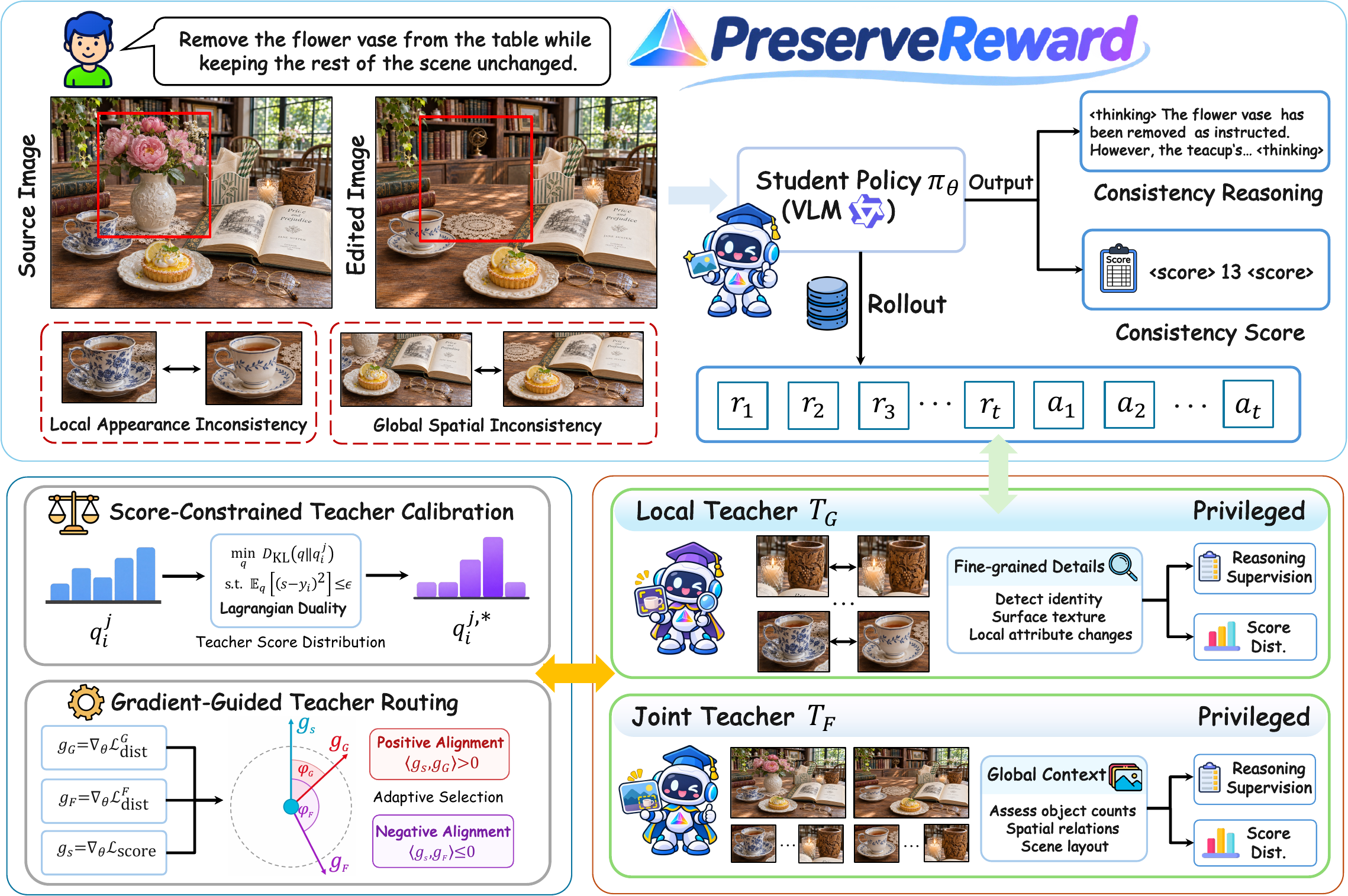}
    \caption{{Overall framework of the proposed PreserveReward.}
    The student generates consistency reasoning and a score from an editing instruction and image pair. Two privileged teachers provide complementary visual supervision through on-policy self-distillation, enhanced by score-constrained calibration and gradient-guided routing.}
    \label{fig:framework}
\end{figure*}

\subsection{Problem Formulation and Overview}
\noindent \textbf{Task Formulation.}
We consider the task of evaluating non-edited subject consistency in instruction-guided image editing. Given a source image $I_i^{\mathrm{src}}$, an edited image $I_i^{\mathrm{edit}}$, and an editing instruction $u_i$, the goal is to assess how faithfully the edited image preserves the source content that should remain unchanged. We denote the training dataset as $\mathcal{D}=\{(x_i,y_i)\}_{i=1}^{N}$, where $x_i=(u_i,I_i^{\mathrm{src}},I_i^{\mathrm{edit}})$ and $y_i$ is the reference preservation target. Our reward model, parameterized by $\theta$, first generates a reasoning trajectory $r_i \sim \pi_\theta(\cdot\mid x_i)$ and then predicts a consistency score from the discrete score space $\mathcal{S}=\{0,\ldots,25\}$.

\noindent \textbf{Score Supervision.}
To align the reward model with human judgments, we adopt a regression-based scoring objective. Let $a_s$ denote the complete answer suffix corresponding to score $s$. We obtain the student's score distribution by normalizing the probabilities of all valid score candidates,
\begin{equation}
p_\theta(s\mid x_i,r_i)
=
\frac{P_\theta(a_s\mid x_i,r_i)}
{\sum_{k\in\mathcal{S}}P_\theta(a_k\mid x_i,r_i)}.
\label{eq:score_prob}
\end{equation}
We then define the Score-MSE loss as the expected normalized squared error between candidate scores and the reference target,
\begin{equation}
\mathcal{L}_{\mathrm{score}}
=
\frac{1}{B}
\sum_{i=1}^{B}
\sum_{s\in\mathcal{S}}
p_\theta(s\mid x_i,r_i)
\left(\frac{s-y_i}{25}\right)^2,
\label{eq:score_mse}
\end{equation}
where $B$ is the batch size. Unlike squared error on the expected score, this objective penalizes errors across the score distribution, directly supervising numerical consistency evaluation.

\noindent \textbf{Training Framework.}
While score supervision aligns predictions with reference targets, it provides limited guidance for visual reasoning. As illustrated in Fig.~\ref{fig:framework}, PreserveReward employs two complementary teacher views for on-policy self-distillation. The local teacher uses matched crops to detect subtle visual changes, while the joint teacher additionally observes the full image pair to assess these changes in global context. Both teachers evaluate the same student-generated trajectories to provide process- and score-level supervision. To improve supervision reliability, we calibrate teacher score distributions using reference targets and apply score-guided gradient routing to select compatible teacher supervision based on the alignment between each teacher's combined distillation gradient and the Score-MSE gradient. At inference, PreserveReward requires only the editing instruction and full image pair.

\subsection{Dual-Teacher On-Policy Self-Distillation}
\noindent \textbf{Privileged Visual Conditioning.}
Evaluating non-edited consistency requires detecting subtle local changes and assessing their validity in the global context. We therefore introduce two privileged teachers with complementary visual evidence. Given the source and edited images, we construct matched crop pairs $\mathcal{G}_i=\{(C_{i,m}^{\mathrm{src}},C_{i,m}^{\mathrm{edit}})\}_{m=1}^{M_i}$, each covering a corresponding region before and after editing. The two teachers are conditioned on:
\begin{align}
x_i^{G} &= (u_i,\mathcal{G}_i),
\label{eq:teacher_local}\\
x_i^{F} &= (u_i,I_i^{\mathrm{src}},I_i^{\mathrm{edit}},\mathcal{G}_i).
\label{eq:teacher_joint}
\end{align}
The local teacher focuses on matched crops to detect subtle changes in subject identity, textures, and visual attributes, preventing fine-grained inconsistencies from being overlooked. The joint teacher additionally incorporates the full image pair to assess object counts, spatial relationships, and scene layout, distinguishing intended edits from unintended changes. They provide complementary supervision for evaluating both local fidelity and global consistency.

\noindent \textbf{On-Policy Self-Distillation.}
To transfer privileged visual knowledge to the student, we perform distillation on reasoning trajectories sampled from the student policy. Given the original input $x_i$, the student generates a reasoning trajectory $r_i\sim\pi_\theta(\cdot\mid x_i)$. Both teachers then evaluate the same trajectory under their respective visual conditions, rather than generating independent responses. At each reasoning position $t$, the student and teacher predictive distributions are given by:
\begin{align}
P_{i,t}^{S} &= P_\theta(\cdot\mid x_i,r_{i,<t}),
\label{eq:student_distribution}\\
P_{i,t}^{j} &= P_{\bar{\theta}}(\cdot\mid x_i^{j},r_{i,<t}),
\quad j\in\{G,F\}.
\label{eq:teacher_distribution}
\end{align}
We align these distributions at the corresponding reasoning positions and minimize their Jensen--Shannon divergence (JSD) \cite{agarwal2024policy}. The process distillation objective for each teacher is defined as:
\begin{equation}
\mathcal{L}_{\mathrm{proc}}^{j}
=
\frac{1}{|\mathcal{B}^{+}|}
\sum_{i\in\mathcal{B}^{+}}
\frac{1}{|\mathcal{A}_i|}
\sum_{t\in\mathcal{A}_i}
D_{\mathrm{JS}}
\left(
\widetilde{P}_{i,t}^{S}
\parallel
\widetilde{P}_{i,t}^{j}
\right),
\label{eq:process_distillation}
\end{equation}
where $\mathcal{A}_i$ denotes valid reasoning positions excluding the final score, and $\mathcal{B}^{+}$ contains samples with nonempty reasoning trajectories. For efficiency, $\widetilde{P}$ represents the distribution over the student's top-$K$ tokens and a residual probability bin, with shared support across teachers. 

\subsection{Score-Constrained Teacher Calibration}
\label{sec:score_calibration}

\noindent \textbf{Teacher Score Calibration.}
Privileged visual information does not guarantee accurate teacher scores, and directly distilling these predictions may propagate scoring bias. We therefore calibrate both teachers' score distributions using reference targets. For each teacher $j\in\{G,F\}$, the score distribution conditioned on the student-generated reasoning $r_i$ is
\begin{equation}
q_i^j(s)
=
\frac{P_{\bar{\theta}}(a_s\mid x_i^j,r_i)}
{\sum_{k\in\mathcal{S}}P_{\bar{\theta}}(a_k\mid x_i^j,r_i)}.
\label{eq:teacher_score_distribution}
\end{equation}
Following Score-MSE, we define the normalized squared error $d_i(s)=((s-y_i)/25)^2$ and its expectation $\mathcal{E}_i(q)=\sum_{s\in\mathcal{S}}q(s)d_i(s)$. To constrain scoring errors while preserving the original teacher distribution, we formulate calibration objective as:
\begin{equation}
\begin{aligned}
q_i^{j,*}
=
\arg\min_{q\in\Delta_{\mathcal{S}}}
&\ D_{\mathrm{KL}}(q\|q_i^j)\\
\mathrm{s.t.}\quad
&\mathcal{E}_i(q)\leq b,
\end{aligned}
\label{eq:score_projection}
\end{equation}
where $\Delta_{\mathcal{S}}$ is the probability simplex over valid scores and $b$ denotes the expected scoring-error budget.

\noindent \textbf{Lagrangian Optimization.}
We solve the constrained projection using the Lagrangian duality:
\begin{equation}
\begin{aligned}
\mathcal{J}(q,\lambda,\nu)
={}& D_{\mathrm{KL}}(q\|q_i^j)
+\lambda(\mathcal{E}_i(q)-b)\\
&+\nu\left(\sum_{s\in\mathcal{S}}q(s)-1\right),
\end{aligned}
\label{eq:calibration_lagrangian}
\end{equation}
where $\lambda\geq0$ and $\nu$ are the multipliers for the error and normalization constraints. Setting $\partial\mathcal{J}/\partial q(s)=0$ yields the exponential reweighting solution:
\begin{equation}
q_{i,\lambda}^{j}(s)
=
\frac{q_i^j(s)\exp(-\lambda d_i(s))}
{\sum_{k\in\mathcal{S}}q_i^j(k)\exp(-\lambda d_i(k))}.
\label{eq:calibrated_score_distribution}
\end{equation}
If $\mathcal{E}_i(q_i^j)\leq b$, the optimal multiplier is $\lambda^*=0$. Otherwise, $\lambda^*$ satisfies $\mathcal{E}_i(q_{i,\lambda^*}^{j})=b$ when the constraint is feasible in the interior. Since $\partial\mathcal{E}_i(q_{i,\lambda}^{j})/\partial\lambda=-\operatorname{Var}_{q_{i,\lambda}^{j}}[d_i(s)]\leq0$, it can be efficiently obtained through one-dimensional search. The calibrated target is then $q_i^{j,*}=q_{i,\lambda^*}^{j}$.

\noindent \textbf{Calibrated Score Distillation.}
We use the calibrated distributions as soft targets to supervise
student score predictions. Let
$p_{\theta,i}=p_\theta(\cdot\mid x_i,r_i)$.
For each teacher $j\in\{G,F\}$, we define:
\begin{equation}
\mathcal{L}_{\mathrm{scoreKD}}^j
=
\frac{\omega_j}{B}
\sum_{i=1}^{B}
m_i^j
D_{\mathrm{JS}}
\left(
p_{\theta,i}
\parallel
\mathrm{sg}(q_i^{j,*})
\right),
\label{eq:calibrated_score_kd}
\end{equation}
where $\omega_j\geq0$ weights the score-distillation loss
of teacher $j$, $m_i^j$ indicates a valid calibration,
and $\mathrm{sg}(\cdot)$ stops gradients through teacher targets. This objective complements the original Score-MSE supervision while mitigating teacher scoring bias. 

\begin{table*}[ht]
\centering
\caption{Pairwise accuracy (\%) for non-edit consistency.
EditReward-Bench covers subject manipulation, appearance modification,
scene restructuring, and advanced editing.
EditReward-Compass covers general editing and reasoning, dynamic manipulation,
multi-image integration, and complex instruction execution.
Bold and underlined values indicate the best and second-best results.}
\label{tab:PreserveRewardbenchmarks}
\small
\setlength{\tabcolsep}{3pt}
\renewcommand{\arraystretch}{1.12}
\resizebox{0.9\textwidth}{!}{%
\begin{tabular}{@{}l*{11}{c}@{}}
\toprule
\multirow{2}{*}{\textbf{Model}}
& \multicolumn{5}{c}{\textbf{EditReward-Bench}}
& \multicolumn{5}{c}{\textbf{EditReward-Compass}}
& \multirow{2}{*}{\textbf{EBench-18K}} \\
\cmidrule(lr){2-6}\cmidrule(lr){7-11}
& Object & Appearance & Scene & Advanced & Avg.
& General & Dynamic & Multi-image & Complex & Avg. & \\
\midrule

\rowcolor{prismopen}
\multicolumn{12}{l}{\textit{Open-source Models}} \\
Qwen3.5-9B
& 37.70 & 41.80 & 37.20 & 40.90 & 39.80
& 35.77 & 47.14 & 51.52 & 36.36 & 39.22 & 56.70 \\
Qwen3.5-35B-A3B
& 38.10 & 40.50 & 62.80 & 45.40 & 44.90
& 49.63 & 60.00 & 51.51 & 60.60 & 52.67 & 61.80 \\
Qwen3.5-397B-A17B
& 46.40 & 60.20 & 59.20 & 70.30 & 60.00
& 69.34 & 69.28 & 72.73 & 57.57 & 68.88 & 72.94 \\
Qwen3.8-27B
& 60.80 & 55.40 & 60.80 & 63.00 & 59.80
& 59.12 & 63.57 & 39.39 & 39.39 & 58.02 & 70.40 \\
Qwen3.8-Flash-Next
& 52.30 & 68.90 & 61.70 & 71.00 & 64.60
& 76.39 & 71.42 & 69.69 & 63.63 & 74.23 & 72.98 \\
MiniMax-M3
& 60.80 & 70.50 & 76.10 & 71.30 & 69.30
& 80.29 & 75.00 & 66.67 & 69.70 & 77.80 & 62.84 \\
GLM-5.3-Flash
& 71.20 & 74.30 & 76.20 & 80.50 & 75.80
& 80.29 & 72.14 & 69.70 & 69.70 & 77.31
& \underline{78.46} \\
Kimi-K3
& 73.30 & 73.80 & \underline{81.10} & 80.70 & 76.90
& 77.37 & 72.14 & 72.73 & 72.73 & 75.69 & 78.05 \\
\midrule

\rowcolor{prismclosed}
\multicolumn{12}{l}{\textit{Proprietary Models}} \\
Gemini-3.1-Flash
& 37.30 & 50.60 & 68.20 & 61.60 & 53.60
& 61.55 & 62.14 & 54.54 & 54.54 & 60.94 & 62.95 \\
Gemini-3.1-Pro
& 68.20 & 73.60 & 63.60 & 81.50 & 73.30
& 77.86 & 65.00 & 66.67 & 63.64 & 73.58 & 66.97 \\
GPT-6-Sol
& 79.90 & 76.00 & 74.70 & 83.30 & 79.00
& \underline{84.18} & 75.00 & \underline{75.76}
& \underline{81.82} & \underline{81.52} & 77.50 \\
GPT-6.1-Sol
& \underline{80.90} & \underline{83.90} & 75.70
& \underline{86.50} & \underline{82.70}
& \textbf{84.42} & 68.57 & 72.73 & 75.76 & 79.74 & 77.94 \\
\midrule

\rowcolor{prismreward}
\multicolumn{12}{l}{\textit{Open-source Reward Models}} \\
EditScore-7B
& 58.50 & 61.20 & 73.50 & 50.30 & 59.10
& 52.80 & 57.86 & 42.42 & 48.48 & 53.16 & 64.37 \\
EditScore-72B
& 52.40 & 62.30 & 69.50 & 54.10 & 58.60
& 57.91 & 65.00 & 33.33 & 57.58 & 58.18 & 64.16 \\
FIRM-8B
& 46.20 & 47.40 & 57.70 & 63.50 & 53.70
& 55.23 & 50.00 & 51.51 & 33.33 & 52.13 & 57.67 \\
RationalRewards-8B-Edit
& 57.10 & 68.60 & 66.00 & 68.10 & 65.40
& 60.82 & 67.14 & 42.42 & 57.57 & 61.10 & 59.58 \\
UnifiedReward-2.0-9B
& 52.20 & 66.20 & 68.30 & 52.20 & 59.00
& 55.23 & 55.00 & 45.45 & 51.51 & 54.45 & 66.76 \\
\midrule

\rowcolor{prismours}
\multicolumn{12}{l}{\textit{Reward Models (Ours)}} \\

\textbf{PreserveReward-9B}
& 78.30 & 81.40 & 79.50 & 82.00 & 80.60
& 79.80 & \underline{80.00} & 72.73
& 69.69 & 78.93 & 76.61 \\
\prismdelta{$\Delta$ vs.\ Qwen3.5-9B}
& \prismdelta{+40.60} & \prismdelta{+39.60}
& \prismdelta{+42.30} & \prismdelta{+41.10}
& \prismdelta{+40.80}
& \prismdelta{+44.03} & \prismdelta{+32.86}
& \prismdelta{+21.21} & \prismdelta{+33.33}
& \prismdelta{+39.71} & \prismdelta{+19.91} \\

\textbf{PreserveReward-27B}
& \textbf{82.90} & \textbf{84.90} & \textbf{82.60}
& \textbf{86.90} & \textbf{84.70}
& 82.72 & \textbf{90.00} & \textbf{84.85}
& \textbf{84.85} & \textbf{84.60} & \textbf{80.73} \\
\prismdelta{$\Delta$ vs.\ Qwen3.8-27B}
& \prismdelta{+22.10} & \prismdelta{+29.50}
& \prismdelta{+21.80} & \prismdelta{+23.90}
& \prismdelta{+24.90}
& \prismdelta{+23.60} & \prismdelta{+26.43}
& \prismdelta{+45.46} & \prismdelta{+45.46}
& \prismdelta{+26.58} & \prismdelta{+10.33} \\
\bottomrule
\end{tabular}%
}

\end{table*}
\subsection{Gradient-Guided Teacher Routing}
\label{sec:gradient_routing}

\noindent \textbf{Teacher Distillation.}
Although score calibration mitigates teacher bias, distillation updates may still conflict with score supervision. As teacher utility varies across editing scenarios, we combine process and calibrated score distillation for adaptive selection,
\begin{equation}
\mathcal{L}_{\mathrm{dist}}^j
=
\mathcal{L}_{\mathrm{proc}}^j
+\alpha\mathcal{L}_{\mathrm{scoreKD}}^j,
\quad j\in\{G,F\},
\label{eq:joint_teacher_distillation}
\end{equation}
\noindent where $\alpha$ balances process and score distillation. Each teacher provides a unified distillation signal combining visual reasoning supervision with calibrated score distributions.

\noindent \textbf{Score-Guided Gradient Feedback.}
We use the Score-MSE gradient computed from reference targets to assess the compatibility of each teacher's distillation signal. For each training batch, we compute:
\begin{equation}
\begin{aligned}
g_s &= \nabla_\theta\mathcal{L}_{\mathrm{score}},\\
g_j &= \nabla_\theta\mathcal{L}_{\mathrm{dist}}^j,
\quad j\in\{G,F\}.
\end{aligned}
\label{eq:routing_gradients}
\end{equation}
All gradients are computed using the same student parameters and fixed student-generated trajectories. Under a first-order approximation, the additional change in scoring loss induced by teacher $j$, relative to score-only optimization, is
\begin{equation}
\Delta_j\mathcal{L}_{\mathrm{score}}
\approx -\eta\lambda_{\mathrm{dist}}
\langle g_s,g_j\rangle,
\label{eq:routing_first_order}
\end{equation}
where $\eta$ is the learning rate and $\lambda_{\mathrm{dist}}>0$ controls the distillation strength. Accordingly, we define the estimated improvement as:
\begin{equation}
R_j=\lambda_{\mathrm{dist}}
\langle g_s,g_j\rangle,
\quad j\in\{G,F\}.
\label{eq:routing_improvement}
\end{equation}
A positive $R_j$ indicates that the teacher's distillation gradient is locally aligned with the scoring objective, while a negative value indicates an optimization conflict. Unlike cosine similarity, this measure accounts for both gradient direction and magnitude.

\noindent \textbf{Adaptive Teacher Routing.}
We adopt a batch-level hard routing strategy that selects the teacher with the highest estimated improvement. The routing decision is defined as:
\begin{equation}
j^*=
\begin{cases}
\varnothing, & R_G<0 \ \land\ R_F<0,\\
\displaystyle\arg\max_{j\in\{G,F\}}R_j,
& \text{otherwise}.
\end{cases}
\label{eq:teacher_routing}
\end{equation}
When both teachers yield conflicting gradients, distillation is disabled for the current batch. Otherwise, the teacher with the larger $R_j$ is selected, with ties resolved in favor of the local teacher. This allows PreserveReward to adapt its supervision to the current training batch while reducing interference with score optimization.

\noindent \textbf{Overall Training Objective.}
Combining reference-score supervision with adaptively selected teacher distillation, the final objective is:
\begin{equation}
\begin{aligned}
\mathcal{L}_{\mathrm{total}}
={} \mathcal{L}_{\mathrm{score}}
+\lambda_{\mathrm{dist}}
\sum_{j\in\{G,F\}}
z_j\mathcal{L}_{\mathrm{dist}}^j,
\end{aligned}
\label{eq:overall_objective}
\end{equation}
where $z_j=\mathbb{I}[j=j^*]$ is a non-differentiable routing indicator. The resulting student gradient is:
\begin{equation}
g_{\mathrm{update}}=
\begin{cases}
g_s, & j^*=\varnothing,\\
g_s+\lambda_{\mathrm{dist}}g_{j^*},
& j^*\in\{G,F\}.
\end{cases}
\label{eq:final_update_gradient}
\end{equation}
The Score-MSE objective is always retained, while at most one teacher contributes joint process and score distillation at each optimization step. 

\section{Experiments}

\subsection{Implementation Details}
\noindent \textbf{Model and Training.}  We initialize the reward models from Qwen3.5-9B \cite{qwen2026qwen35} and Qwen3.8-27B \cite{qwen2026qwen38_27b} and train them for one epoch using veRL~\cite{sheng2025hybridflow}. We use AdamW with a learning rate of $2 \times 10^{-6}$, 10 warm-up steps, and a global batch of 96 prompts, sampling one response per prompt. We set the distillation loss weight to $0.25$, the teacher EMA decay to $0.95$, and the score calibration error budget to $0.01$. For downstream reinforcement learning, we post-train Qwen-Image-2.1~\cite{qwen2026qwenimage21} using the learned preservation scores as a reward signal. We use GRPO in Flow-Factory \cite{ping2026flowfactory} to train rank-64 LoRA adapters at a learning rate of $3 \times 10^{-4}$, sampling 16 candidate edits per instruction.

\noindent \textbf{Training Data Construction.}  We construct our training data from EditScore~\cite{luo2026editscore} and EditHF-1M~\cite{xu2026edithf1mmillionscalerichhuman}. We employ Qwen3.8-27B~\cite{qwen2026qwen38_27b} to ground consistency-relevant targets and extract matched crops as privileged visual evidence for the teachers. Qwen3.5-397B~\cite{qwen2026qwen35} assesses grounding quality to filter inaccurate localization. The final set contains 12,231 samples: 10,000 from EditScore and 2,231 from EditHF-1M.

\begin{table*}[ht]
\centering
\caption{
Results after reinforcement learning with different reward models
on three image-editing benchmarks, using Qwen-Image-2.1
as the policy model.
We report selected subcategories alongside overall scores.
\textit{Base} denotes the generator without RL fine-tuning.
}
\label{tab:PreserveRewardrl}
\small
\setlength{\tabcolsep}{4pt}
\renewcommand{\arraystretch}{1.16}

\resizebox{\textwidth}{!}{%
\begin{tabular}{@{}l*{14}{c}@{}}
\toprule

\multirow{2}{*}{\textbf{Reward Model}}
& \multicolumn{5}{c}{\textbf{GEdit-Bench-v1}}
& \multicolumn{4}{c}{\textbf{GEdit-Bench-v2}}
& \multicolumn{5}{c}{\textbf{KRIS-Bench}} \\

\cmidrule(lr){2-6}
\cmidrule(lr){7-10}
\cmidrule(lr){11-15}

&
\makecell[c]{Color\\Alteration} &
\makecell[c]{Material\\Alteration} &
\makecell[c]{Style\\Change} &
\makecell[c]{Subject\\Removal} &
\makecell[c]{Overall} &
\makecell[c]{Character\\Reference} &
\makecell[c]{Subject\\Addition} &
\makecell[c]{Text\\Editing} &
\makecell[c]{Overall} &
\makecell[c]{Count\\Change} &
\makecell[c]{Abstract\\Reasoning} &
\makecell[c]{Practical\\Knowledge} &
\makecell[c]{Science} &
\makecell[c]{Overall} \\

\midrule

\textit{Base}
& {9.24} & 8.67 & 8.75 & 8.79 & 8.86
& 8.63 & 9.61 & 9.30 & 8.83
& 3.16 & 3.73 & 3.94 & 3.77 & 3.67 \\

EditScore-7B
& 9.23 & 8.69 & \underline{8.92} & {8.89} & {8.93}
& 8.65 & 9.60 & 9.29 & 8.92
& 3.26 & 3.37 & 3.76 & 3.78 & 3.62 \\

EditScore-72B
& 9.00 & 8.59 & 8.86 & 8.71 & 8.73
& 8.40 & 9.60 & 9.01 & 8.70
& 3.00 & 3.25 & 3.44 & 3.31 & 3.28 \\

FIRM-8B
& 9.19 & 8.60 & 8.77 & 8.75 & 8.75
& 8.55 & 9.63 & 9.11 & 8.61
& 3.12 & 3.51 & 3.82 & 3.45 & 3.41 \\

RationalRewards-8B-Edit
& 8.89 & 8.36 & 8.64 & 8.04 & 8.47
& 8.23 & 8.95 & 8.57 & 8.29
& 3.16 & 3.23 & 3.20 & 3.45 & 3.33 \\

UnifiedReward-2.0-9B
& 9.04 & 8.44 & 8.82 & 7.45 & 8.57
& 7.88 & 8.53 & 8.53 & 8.21
& 3.50 & 3.92 & 4.02 & 3.90 & 3.85 \\

\textbf{PreserveReward-9B}
& \underline{9.28} & \textbf{8.85} & {8.88}
& \textbf{8.97} & \underline{8.95}
& \underline{8.73} & \underline{9.64}
& \underline{9.40} & \underline{8.94}
& \underline{3.64} & \underline{4.01}
& \underline{4.10} & \underline{3.98}
& \underline{3.86} \\

\textbf{PreserveReward-27B}
& \textbf{9.39} & \underline{8.83} & \textbf{8.98}
& \underline{8.96} & \textbf{8.99}
& \textbf{8.80} & \textbf{9.83}
& \textbf{9.41} & \textbf{9.02}
& \textbf{3.90} & \textbf{4.15}
& \textbf{4.20} & \textbf{4.16}
& \textbf{4.07} \\

\bottomrule
\end{tabular}%
}
\end{table*}
\begin{table}[t]
\centering
\caption[Ablations on EditReward-Bench]{{Ablations on EditReward-Bench.}
Pairwise accuracy (\%).
Obj., App., and Adv. denote Object, Appearance, and Advanced.
}
\label{tab:PreserveRewardablation}
\small
\setlength{\tabcolsep}{3pt}
\renewcommand{\arraystretch}{1.15}

\resizebox{0.88\columnwidth}{!}{%
\begin{tabular}{
@{}lcccc>{\columncolor[gray]{0.96}[\tabcolsep][0pt]}c@{}
}
\toprule
\textbf{Variant}
& \textbf{Obj.}
& \textbf{App.}
& \textbf{Scene}
& \textbf{Adv.}
& \textbf{Avg.} \\
\midrule
\textit{Base}
& 37.70 & 41.80 & 37.20 & 40.90 & 39.80 \\
Score-MSE only
& 54.30 & 75.90 & 64.80 & 69.70 & 67.30 \\
Local Teacher only
& 64.00 & 72.50 & 78.20 & 75.70 & 72.40 \\
Joint Teacher only
& 64.90 & 78.50 & 75.00 & 74.40 & 73.50 \\
w/o Dist. Calibration
& 64.00 & 75.00 & 79.80 & 78.30 & 74.20 \\
w/o Gradient Routing
& 67.10 & 78.40 & 79.30 & 78.20 & 75.90 \\

\midrule
\textbf{PreserveReward-9B}
& \textbf{78.30} & \textbf{81.40} & \textbf{79.50} & \textbf{82.00} & \textbf{80.60} \\
\bottomrule
\end{tabular}%
}
\end{table}
\noindent \textbf{Benchmarks.}  We evaluate visual consistency at both the reward model and image editor levels. To assess agreement with human preferences on visual consistency, we use the consistency subsets of EditReward-Bench~\cite{luo2026editscore} and EditReward-Compass~\cite{bai2026edit}, together with the high-level testset of EBench-18K~\cite{xu2025lmm4edit}. For EBench-18K, we derive pairwise preferences from human preservation annotations for edits that share the same source image and instruction. To assess whether RL with our reward improves the preservation of subjects and regions outside the scope of the requested edit, we compare each editor before and after RL on GEdit-Bench v1~\cite{liu2025step1x}, GEditBench v2~\cite{jiang2026geditbench} and KRIS-Bench~\cite{wu2026kris}. We report results for selected subcategories alongside overall scores.

\noindent \textbf{Baselines.} We compare PreserveReward against specialized reward models and general-purpose vision-language models (VLMs) on preservation preference benchmarks. For open-source reward models, we include EditScore-7B and EditScore-72B~\cite{luo2026editscore}, FIRM-8B~\cite{zhao2026trust}, RationalRewards-8B-Edit \cite{wang2026rationalrewards}, and UnifiedReward-2.0-9B \cite{unifiedreward}. For general-purpose open-source VLMs, we evaluate the Qwen3.5 series~\cite{qwen2026qwen35} at three scales (9B, 35B-A3B, and 397B-A17B), alongside Qwen3.8-27B~\cite{qwen2026qwen38_27b} and Qwen3.8-Flash-Next~\cite{qiu2026design}. We further include large-scale open-source models such as MiniMax-M3~\cite{minimax2026m3}, GLM-5.3-Flash~\cite{zai2026glm53flash}, and Kimi-K3~\cite{team2026kimi}. The advanced proprietary baselines include Gemini-3.1-Flash~\cite{google2026gemini31flashlite}, Gemini-3.1-Pro~\cite{google2026gemini31pro}, GPT-6-Sol~\cite{openai2026gpt6sol}, and GPT-6.1-Sol~\cite{openai2026gpt61sol}.

\noindent \textbf{Evaluation Metrics.} For direct reward model evaluation, we report pairwise accuracy against human preservation judgments. We construct reference preference pairs by comparing human-annotated preservation scores. A prediction is correct if the preferred edit scores higher or, for a reference tie, the absolute score difference is below $10^{-5}$. To measure how well image editors preserve content outside the scope of the requested edit, we use GPT-6.1-Sol to compare the source image with the outputs of both the base and RL-trained editors. For GEdit-Bench v1 and GEditBench v2, we use the same VIEScore-derived over-editing score ($0$--$10$), where higher scores indicate fewer unintended changes. For KRIS-Bench, we use the single-image consistency score ($1$--$5$), which measures how well the output retains source attributes that the instruction does not ask to change.

\subsection{Main Results}
\noindent \textbf{Human Preference Alignment.} As shown in Tab.~\ref{tab:PreserveRewardbenchmarks}, PreserveReward-27B achieves the highest average pairwise accuracy across all three benchmarks, reaching 84.70\% on EditReward-Bench, 84.60\% on EditReward-Compass, and 80.73\% on EBench-18K, with the latter two serving as out-of-distribution evaluations. It consistently outperforms existing reward models and competitive general-purpose VLMs, including substantially larger models. Notably, PreserveReward-9B also demonstrates strong performance, achieving accuracies of 80.60\%, 78.93\%, and 76.61\%, respectively, surpassing all evaluated open-source reward models. These results demonstrate the effectiveness of PreserveReward in aligning consistency evaluation with human preference.

\noindent \textbf{Reinforcement Fine-Tuning.}
As shown in Tab.~\ref{tab:PreserveRewardrl}, PreserveReward-27B improves Qwen-Image-2.1 across all three image-editing benchmarks, achieving the best overall scores on both GEdit-Bench-v1 and GEdit-Bench-v2. Notable gains in human editing and subject replacement further highlight its ability to preserve visual content under diverse editing operations. Meanwhile, PreserveReward-27B improves the overall KRIS-Bench score from 3.67 to 4.07, outperforming all compared reward models with consistent gains across all four categories. These results validate the effectiveness of PreserveReward in guiding RL fine-tuning toward more faithful content preservation.

\begin{figure}[!t]
    \centering
    \includegraphics[width=\linewidth]{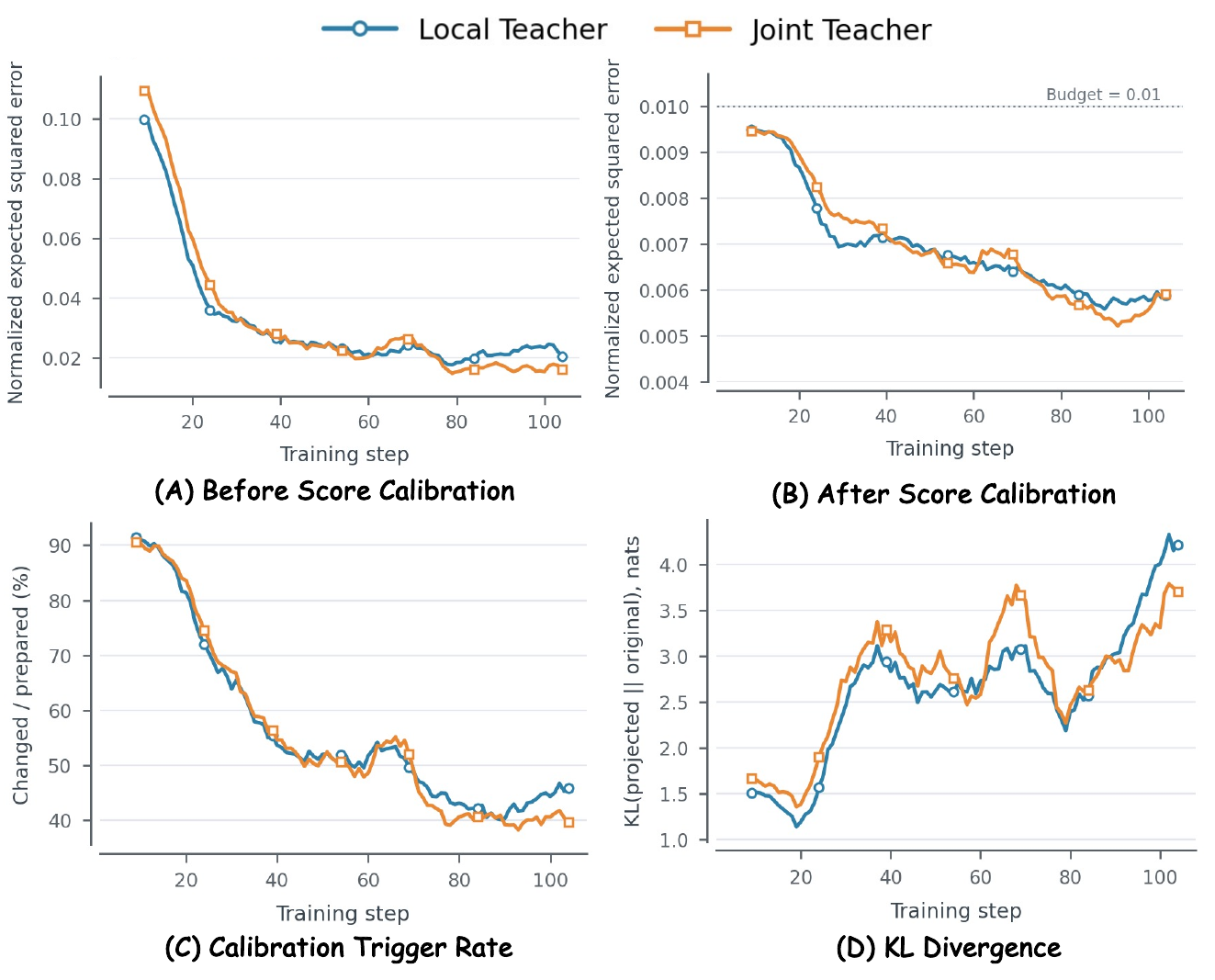}
    \caption{{Dynamics of score-constrained teacher calibration.}
    (A) Expected scoring error before calibration.
    (B) Expected scoring error after calibration, with the error budget of 0.01.
    (C) Fraction of teacher distributions requiring calibration.
    (D) KL divergence between calibrated and original teacher distributions.}
    \label{fig:calibration_analysis}
\end{figure}
\begin{figure}[!t]
    \centering
    \includegraphics[width=\linewidth]
    {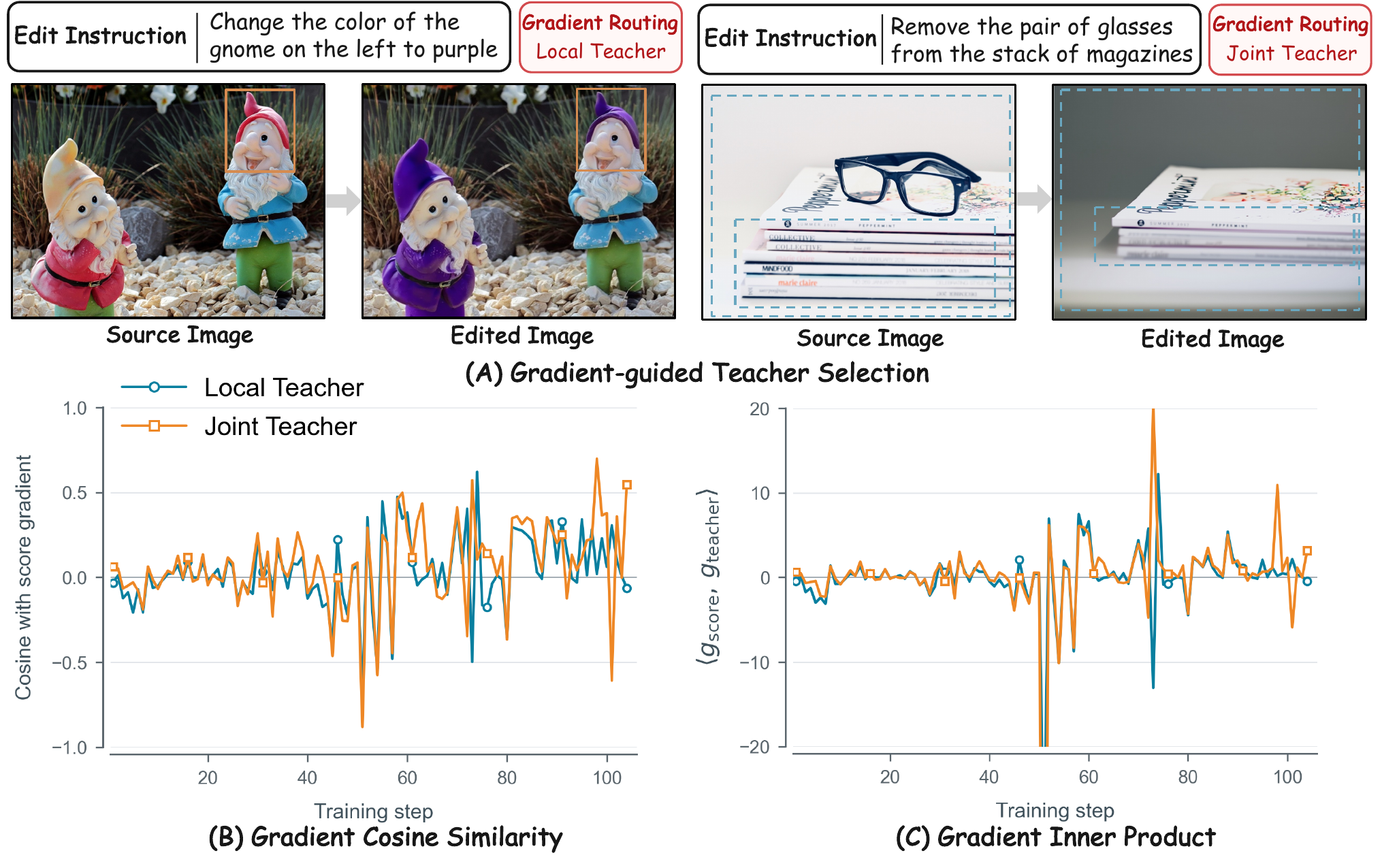}
    \caption{
        {Dynamics of gradient-guided teacher selection.}
        (A) Examples of unintended local appearance and global spatial changes in non-target content.
        (B) Cosine similarity between the score-loss gradient and each teacher's distillation gradient.
        (C) Corresponding gradient inner products during training.
    }
    \label{fig:gradient_guided_teacher_selection}
\end{figure}
\subsection{Ablation Study}
Tab.~\ref{tab:PreserveRewardablation} presents ablations on EditReward-Bench. Score-MSE substantially improves alignment with human consistency judgments over the base model, while either teacher provides further gains in overall accuracy. The local teacher performs better on scene and advanced editing, whereas the joint teacher is stronger on appearance modification. Combining both teachers without routing further improves overall accuracy, supporting the complementarity of local details and global context. Removing score-constrained calibration reduces performance, highlighting the importance of reliable teacher score targets. Gradient-guided routing provides additional gains, particularly on object manipulation and advanced editing, supporting the selection of distillation supervision compatible with the scoring objective. The full model achieves the best overall accuracy, demonstrating the combined benefits of complementary visual supervision, teacher calibration, and adaptive routing.

\begin{figure}[!t]
    \centering
    \includegraphics[width=\linewidth]{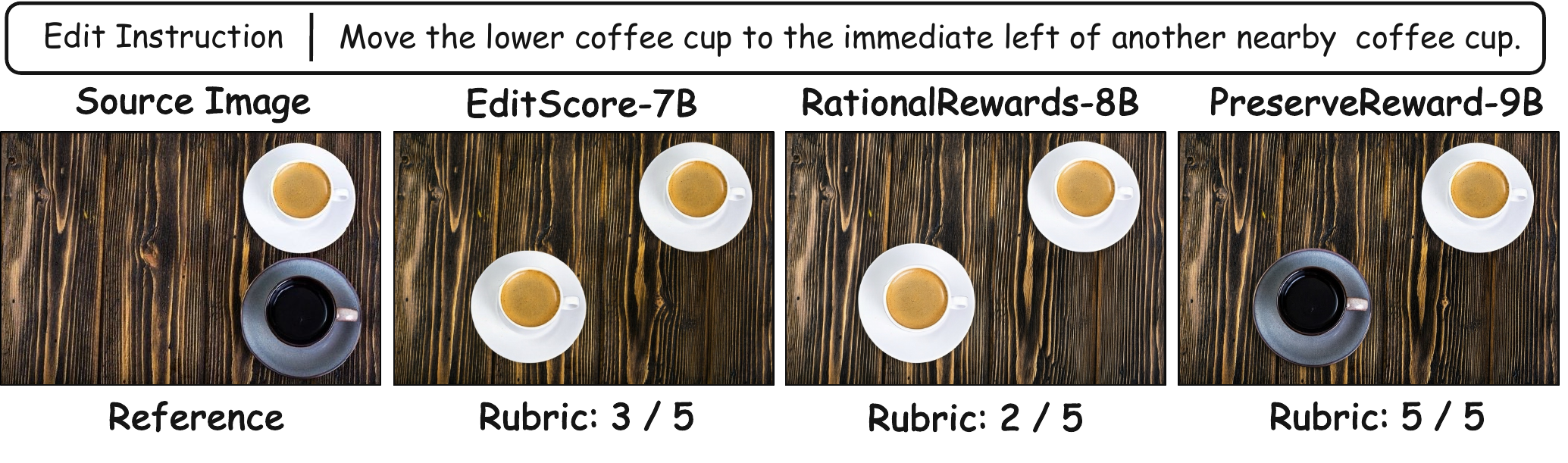}
    \caption{Qualitative comparison after RL fine-tuning with different reward models.}
    \label{fig:consistency_optimization}
\end{figure}

\subsection{Qualitative Results}
\noindent \textbf{Teacher Calibration.} Fig.~\ref{fig:calibration_analysis} shows that calibration consistently constrains both teachers' expected scoring errors within the predefined budget. The calibration trigger rate and KL divergence further characterize the frequency and magnitude of distributional adjustments, demonstrating effective error control while preserving the original teacher distributions.

\noindent \textbf{Gradient-Guided Teacher Routing.}
Fig.~\ref{fig:gradient_guided_teacher_selection} illustrates teacher selection and gradient dynamics during training. As shown in (A), the local teacher is selected for localized appearance changes, while the joint teacher is favored when broader scene context is required. (B) and (C) show the cosine similarity and inner product between each teacher's distillation gradient and the Score-MSE gradient. Guided by these signals, gradient routing dynamically selects teacher supervision compatible with the reference scoring objective, mitigating optimization conflicts.

\noindent \textbf{Content Preservation after RL.} Fig.~\ref{fig:consistency_optimization} presents a qualitative comparison after RL fine-tuning with different reward models. Compared with the baselines, PreserveReward better preserves non-edited content while accurately executing the editing instruction, reducing unintended visual changes. These results further demonstrate its effectiveness in guiding RL toward more faithful image editing.

\section{Conclusion}
In this work, we presented PreserveReward, a consistency reward model for instruction-guided image editing based on dual-teacher on-policy self-distillation. By leveraging privileged local and global visual evidence, PreserveReward enhances the assessment of subtle visual changes and their contextual validity. To improve supervision reliability, we introduce score-constrained teacher calibration via minimum-KL projection, mitigating teacher scoring bias while preserving distributional information. We further employ score-guided gradient routing to adaptively select teacher supervision aligned with the reference scoring objective, thereby reducing optimization conflicts. Extensive experiments demonstrate the effectiveness of PreserveReward in accurately evaluating attribute preservation and its strong alignment with human preference.

{
    \small
    \bibliographystyle{ieeenat_fullname}
    \bibliography{main}
}

\end{document}